\def\year{2022}\relax
\documentclass[letterpaper]{article} 
\usepackage{aaai22}  
\usepackage{times}  
\usepackage{helvet}  
\usepackage{courier}  
\usepackage[hyphens]{url}  
\usepackage{graphicx} 
\usepackage{natbib,booktabs}  
\usepackage{caption} 
\DeclareCaptionStyle{ruled}{labelfont=normalfont,labelsep=colon,strut=off} 
\usepackage{algorithm}
\usepackage{algorithmic}
\usepackage{amssymb}
\usepackage{amsmath}
\usepackage{mathtools}
\usepackage{caption}
\usepackage{subcaption}
\usepackage{graphicx}
\usepackage{tikz}
\usetikzlibrary{positioning}
\newcommand{\ze}{\boldsymbol{z}^e}
\newcommand{\zg}{\boldsymbol{z}^g}
\newcommand{\ye}{\boldsymbol{y}^e}
\newcommand{\yg}{\boldsymbol{y}^g}
\newcommand{\xe}{\boldsymbol{x}^e}
\newcommand{\xg}{\boldsymbol{x}^g}
\newcommand{\pe}{\boldsymbol{p}}
\newcommand{\fg}{\boldsymbol{f}}
\newcommand{\Gtwo}{\boldsymbol{G}_2} 
\newcommand{\Eone}{\boldsymbol{E}_1}
\newcommand{\Etwo}{\boldsymbol{E}_2}
\newcommand{\CEone}{\boldsymbol{c}^e_1}
\newcommand{\CEtwo}{\boldsymbol{c}^e_2}
\newcommand{\CEthree}{\boldsymbol{c}^e_3}
\newcommand{\CGone}{\boldsymbol{c}^g_1}
\newcommand{\CGtwo}{\boldsymbol{c}^g_2}
\newcommand{\CGthree}{\boldsymbol{c}^g_3}
\newcommand{\bEone}{\boldsymbol{b}^e_1}
\newcommand{\bEtwo}{\boldsymbol{b}^e_2}
\newcommand{\bGone}{\boldsymbol{b}^g_1}

\newcommand{\Ae}{\boldsymbol{A}^e}
\newcommand{\Be}{\boldsymbol{B}^e}
\newcommand{\De}{\boldsymbol{D}^e}
\newcommand{\Ag}{\boldsymbol{A}^g}
\newcommand{\Bg}{\boldsymbol{B}^g}
\newcommand{\Dg}{\boldsymbol{D}^g}
\newcommand{\He}{\boldsymbol{H}^e}

\pdfoutput=1

\usepackage{newfloat}
\usepackage{listings}
\floatstyle{ruled}
\newfloat{listing}{tb}{lst}{}
\floatname{listing}{Listing}
\usepackage{xcolor,arydshln}

\title{Graph Representation Learning for Energy Demand Data: \\
Application to Joint Energy System Planning under Emissions Constraints}
\author{
   Aron Brenner \equalcontrib\textsuperscript{\rm 1}, Rahman Khorramfar\equalcontrib\textsuperscript{\rm 2},  Dharik Mallapragada\textsuperscript{\rm 3}, Saurabh Amin\textsuperscript{\rm 4}\\}
\affiliations{
     \textsuperscript{\rm 1,4 }Civil and Environmental Engineering (CEE) and Laboratory for Information $\&$ Decision Systems (LIDS) \\
     \textsuperscript{\rm 2} MIT Energy Initiative (MITEI) and Laboratory for Information $\&$ Decision Systems (LIDS)\\
     \textsuperscript{\rm 3 } MIT Energy Initiative (MITEI)\\
   \{abrenner, khorram, dharik, amins\}@mit.edu
    \\

}

\begin{document}
\maketitle
\begin{abstract}
A rapid transformation of current electric power and natural gas (NG) infrastructure is imperative to meet the mid-century goal of CO$_2$ emissions reduction requires. This necessitates a long-term planning of the joint power-NG system under representative demand and supply patterns, operational constraints, and policy considerations. Our work is motivated by the computational and practical challenges associated with solving the generation and transmission expansion problem (GTEP) for joint planning of power-NG systems. Specifically, we focus on efficiently extracting a set of representative days from power and NG data in respective networks and using this set to reduce the computational burden required to solve the GTEP. We propose a Graph Autoencoder for Multiple time resolution Energy Systems (GAMES) to capture the spatio-temporal demand patterns in interdependent networks and account for differences in the temporal resolution of available data. The resulting embeddings are used in a clustering algorithm to select representative days. We evaluate the effectiveness of our approach in solving a GTEP formulation calibrated for the joint power-NG system in New England. This formulation accounts for the physical interdependencies between power and NG systems, including the joint emissions constraint. Our results show that the set of representative days obtained from GAMES not only allows us to tractably solve the GTEP formulation, but also achieves a lower cost of implementing the joint planning decisions.
\end{abstract}

\section{Introduction}

One of the most significant societal challenges that we currently face is to transition to a reliable, low-carbon, and sustainable energy system as soon as possible, and to meet the mid-century goal of limiting global warming below 2$^\circ$C \cite{ParisAgreement2015,GielenEtal2019}. This requires a significant use of renewable energy resources and well-planned integration of various energy vectors, including emerging clean energy sources such as hydrogen and other renewable energy sources. Our work is motivated by the enormous potential of machine learning (ML) models in promoting sustainable energy systems. In particular, we focus on ML modeling for extracting a set of \textit{representative days} from heterogeneous demand data associated with real-world electric power and natural gas (NG) systems, and using this set for joint power-NG network planning under emissions constraints. In doing so, we leverage ML-extracted representative days to tractably solve an optimization problem that determines a capacity and network expansion plan for regional-scale energy systems such as that of New England. 

Broadly speaking, our work addresses several {practical and computational challenges} associated with capacity expansion models (CEMs) for decarbonization of interdependent power-NG infrastructures. Classical examples of such models include the generation expansion problem (GEP) and generation and transmission expansion problem (GTEP), both of which are well-studied in the context of power systems \cite{LiEtal2022-EJOR,HeEtal-survey2018}. Our optimization model is a GTEP that determines the optimal location and timing of generation units, transmission lines, and pipelines  to meet future energy demands under a range of operational and policy constraints such as joint emission constraints.
In our work, we extend the model to include two main interdependencies between power and NG systems. The first interdependency captures the increasing role of gas-fired power plants in the generation mix of electricity production \cite{eiaWebsite2021,HeEtal-survey2018}. The second interdependency reflects the \textit{joint} emission of CO$_2$ in both systems. 

The key \emph{computational challenge} in solving the GTEP arises from the fact that it links long-term investment decisions (e.g. capacity and network expansion) to short-term operational ones (e.g. unit commitment, power production, and energy storage). The former decisions have a planning horizon of 10-30 years with yearly granularity, while the latter usually require hourly or sub-hourly resolution. Under reasonable assumptions, we can express the GTEP as a large-scale mixed-integer linear program (MILP), but current literature has limited success in tractably solving these problems to an adequate level of spatial and temporal resolution. In our case, the computational difficulty in solving the GTEP increases further because we model both power and NG networks. Thus, taking into account (projected) demand information on a day-to-day basis becomes prohibitively expensive from a computational viewpoint.
In the classical GTEP problems for power systems, the computational challenge is addressed by aggregating power system nodes (buses) within a geographical neighborhood (power zone) to a single node~\cite{LiEtal2022-EJOR} and by solving the GTEP for a set of representative days~\cite{HoffmannEtal2020-survey}. Crucially, the set of representative days needs to capture demand and supply patterns. To the best of our knowledge, the notion of representative days has not been clearly defined and developed in the context of joint power-NG planning problem – this is where we leverage our graph representation learning approach.

\begin{figure*}[hbtp]
    \centering
    \includegraphics[width=\textwidth]{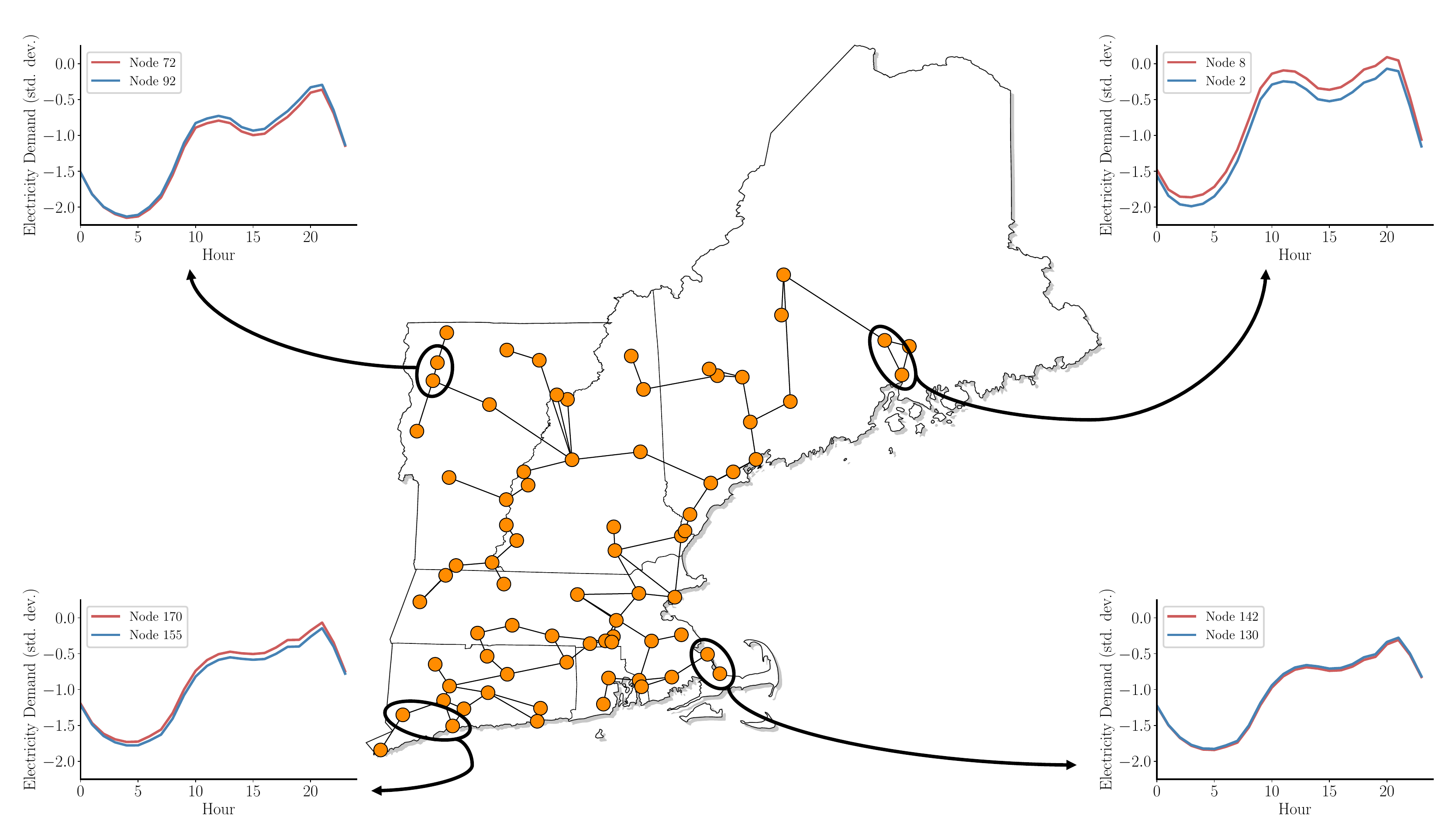}
    \caption{Adjacent nodes in the power network demonstrate similar variations in demand over the course of the day. These spatial dependencies are modeled explicitly by graph convolutional layers in the GAMES architecture.}
    \label{fig:diagram}
\end{figure*}

Our work also addresses the \emph{practical issues} arising from coarse data availability from the NG network. Firstly, we do not have access to the detailed connectivity and transmission information in the NG network while this information is readily available for the power network. Secondly, power systems typically collect demand and generation data at a fine temporal resolution (hourly or less), but this data is usually not publicly accessible for NG systems. These issues thus require us to (a) formulate network constraints based on loosely specified information on power and NG node connectivity and (b) develop an approach to leverage demand and supply data from the power system with demand data of NG system despite their different temporal resolutions. 

We address the aforementioned challenges by developing a \emph{graph representation learning approach} that captures the physical interdependencies between power and NG networks, and also handles the different granularity of data at each network. We consider demand data for both systems, and consider capacity factor (CF) data for solar and wind plants to reflect the supply pattern in the renewable-dominated future grid. 
We utilize graph convolutions to capture the network interactions both within and across power and NG networks, and adopt an autoencoder architecture with tuneable reconstruction losses for the respective demand and CF data. We demonstrate that the resulting \textbf{Graph Autoencoder for Multiple time resolution Energy Systems (GAMES)} model is ideally suited to handle embedding the spatio-temporal patterns in power and NG demand as well as wind and solar CF data into a lower-dimensional representation, which can be readily clustered to extract the set of representative days.
Furthermore, our approach to computing the set of representative days can also enable an accurate estimation of the trade-off between costs (both investment and operational) and joint emissions from power and NG systems.\footnote{We believe this capability can have a significant societal impact by lowering the barriers to investment in renewable energy resources and alleviating reliability concerns in a low-carbon energy system.} 

Previous studies for selecting representative days propose variants of  k-means \cite{MallapragadaEtal2018,LiEtal2022,TeichgraeberBrandt2019,BarbarMallapragada2022},  k-medoids \cite{ScottEtal2019,TeichgraeberBrandt2019}, and hierarchical clustering \cite{LiuEtal2017,TeichgraeberBrandt2019}.
The distance matrices used in clustering algorithms for most previous works are constructed based on a set of time series inputs such as load data and variable renewable energies (VRE) capacity factors \cite{LiEtal2022-EJOR,HoffmannEtal2020-survey}. Notably, these approaches neither account for demand data with multiple time resolutions nor account for network interdependencies. Hence, they cannot be readily extended to address the task of extracting representative days for joint power-NG systems – an aspect that is crucial for realism and tractability in joint planning optimization models for decarbonizing these systems. We believe that our GAMES model addresses these challenges and provides a promising path to better extract representative days in interdependent power and NG systems.





\section{Graph Convolutional Autoencoder Approach}\label{sec:sol}
In this section, we describe the \textit{Graph Autoencoder for Multiple time resolution Energy Systems} (GAMES) model, a simple graph autoencoder with linear graph convolutions. We argue that this architecture efficiently captures spatio-temporal demand patterns in power and NG systems. 


\subsection{Autoencoders}
To begin with, we note that direct use of clustering algorithms to identify representative days for any large-scale energy system is prone to the ``curse of dimensionality'' due to the high dimensionality of time series data. In such settings, it is desirable to first extract low-dimensional and denoised representations of the data before clustering \cite{parsons2004}. To identify a set of representative days, we choose to utilize a state-of-the-art autoencoder architecture for learning low-dimensional embeddings for power-NG systems (that have different time resolutions) prior to clustering.

Given a high-dimensional input such as a time series of graph signals, $X \in \mathbb{R}^{p}$, an autencoder can be trained to jointly learn an encoder, $g: \mathbb{R}^p \to \mathbb{R}^k$, and a decoder, $f: \mathbb{R}^k \to \mathbb{R}^p$ that minimize the reconstruction loss function $\|X - \hat{X}\|_2^2$, where $\hat{X}=f(g(X))$ is the reconstructed signal. Here, $k \ll p$ denotes the dimension of the learned latent space.

\begin{table}[H]
    \centering
    \begin{tabular}{|c|c|c|c|} 
        \hline
        \textbf{Variable} & \textbf{Interpretation} & \textbf{Granularity} & \textbf{Nodes} \\
        \hline
        $X_\mathrm{E}$ & Electricity & Hourly & 188 \\ 
        \hline
        $X_\mathrm{W}$ & Wind & Hourly & 188 \\
        \hline
        $X_\mathrm{S}$ & Solar & Hourly & 188 \\
        \hline
        $X_\mathrm{G}$ & Natural Gas & Daily & 18 \\
        \hline
    \end{tabular}
    \caption{Notation for input variables.}
    \label{tab:notation}
\end{table}

We denote by $X_\mathrm{E} \in \mathbb{R}^{d \times n_\mathrm{E} \times t_\mathrm{E}}$ the data tensor of electricity demands over all days $d$, nodes $n_\mathrm{E}$, and times $t_\mathrm{E}$. Similarly, we denote the natural gas data tensor by $X_\mathrm{G}\in \mathbb{R}^{d \times n_\mathrm{G} \times t_\mathrm{G}}$, the wind capacity factor tensor by $X_\mathrm{W} \in \mathbb{R}^{d \times n_\mathrm{W} \times t_\mathrm{W}}$, and the solar capacity factor data tensor by $X_\mathrm{S} \in \mathbb{R}^{d \times n_\mathrm{S} \times t_\mathrm{S}}$ (see Table \ref{tab:notation}). Because the GTEP considers different associated costs for investment and operational decisions related to power, NG, wind, and solar, we introduce hyperparameters $\alpha_G,\alpha_W,\alpha_S$ in the autoencoder objective function to tune the trade-off between the multiple reconstruction losses. This parameter reflects the contribution of each system towards the total cost. For example, if the NG system cost is twice the power system cost, then higher values of $\alpha_G$ ensure that the reconstruction cost is penalized more when deviating from the data of the NG system. This gives us the following loss function:
\begin{align*}
    \sum_{i=1}^d\Big(\frac{1}{dn_\mathrm{E}t_\mathrm{E}}\|X_\mathrm{E}^{(i)} - \hat{X}_\mathrm{E}^{(i)}\|_F^2 + \frac{\alpha_G}{dn_\mathrm{G}t_\mathrm{G}}\|X_\mathrm{G}^{(i)} - \hat{X}_\mathrm{G}^{(i)}\|_F^2 \\
    + \frac{\alpha_W}{dn_\mathrm{W}t_\mathrm{W}}\|X_\mathrm{W}^{(i)} - \hat{X}_\mathrm{W}^{(i)}\|_F^2 + \frac{\alpha_S}{dn_\mathrm{S}t_\mathrm{S}}\|X_\mathrm{S}^{(i)} - \hat{X}_\mathrm{S}^{(i)}\|_F^2 \Big),
\end{align*}
where $\|\cdot\|_F$ denotes the Frobenius norm.

In our case study, we set $\alpha_G=2, \alpha_S=0.5, \alpha_W=0.5$. However, we note that it is possible to choose the hyperparameters by evaluating the downstream GTEP objective for different values. Specifically, this can be performed using a grid search in which the quality of a combination of hyperparameters $\{\alpha_G,\alpha_W,\alpha_S\}$ is measured by GTEP objective costs given by solving the optimization model rather than the GAMES validation loss directly.

\subsection{Graph Representation Learning}
Next, we provide a brief introduction to modeling with graph convolutional networks (GCNs).

\subsubsection{Preliminaries}
We encode the network topology with the binary adjacency\footnote{Ideally, one should construct an affinity matrix $A$ with a Gaussian kernel such that $A_{ij} = \exp{\left(-\frac{\mathrm{dist}(i,j)^2}{\sigma^2}\right)}$ for all edges $(i,j)$, where $\mathrm{dist}(i,j)$ denotes the distance of edge $(i,j)$ and $\sigma$ denotes the standard deviation of distances in the network \cite{shuman2012}. Since we do not have access to edge distance data in our case study, we proceed with the binary adjacency matrix.} matrix $A$, which we construct such that
\begin{align*}
A_{ij} = 
    \begin{cases} 
 0 & (i,j) \notin \mathcal{E} \\
 1 & (i,j) \in \mathcal{E}.
    \end{cases}
\end{align*}
We also construct the diagonal degree matrix $D$ such that $D_{ii} = \sum_{j} A_{ij}$.

\subsubsection{Graph Convolutions}
Our graph autoencoder approach follows \cite{kipf2017} in utilizing \textit{Chebyshev convolutional filters}, which approximate spectral convolutions to learn node embeddings as weighted local averages of embeddings of adjacent nodes. This is ideal for learning low-dimensional embeddings of energy networks as neighborhoods of nodes typically exhibit similar energy demands patterns and can thus be represented jointly. Chebyshev filters operate on the ``renormalized'' graph Laplacian $\tilde{L} = \tilde{D}^{-\frac{1}{2}}\tilde{A}\tilde{D}^{-\frac{1}{2}}$, where $\tilde{D} = I + D$ and $\tilde{A} = I + A$, and perform a form of Laplacian smoothing \cite{li2018, taubin1995}. We initialize $H^{(0)} = X$ and apply convolutional filters to learn subsequent node embeddings as follows:
\begin{align*}
    H^{(l+1)} = \sigma (\tilde{L} H^{(l)} \Theta^{(l)}),
\end{align*}
where $\Theta^{(l)}$ is a trainable weight matrix and $H^{(l)}$ is a matrix of node embeddings in layer $l$. $\sigma(\cdot)$ is typically a nonlinear activation function, such as $\mathrm{ReLU}$ or $\mathrm{tanh}$.

In each layer, GCNs aggregate features from the immediate neighborhood of each node. Deep GCNs stack multiple layers with nonlinear activations to learn node embeddings as nonlinear functions of both local and global node features. In contrast, \cite{salha2019} propose a simpler graph autoencoder model, which they demonstrate to have competitive performances with multilayer GCNs on standard benchmark datasets despite being limited to linear first-order interactions. Shallow neural architectures are also better suited for settings where data is scarce. This is particularly significant in modeling energy systems whose data may only be available for a few historical years. 
Indeed, we find this simpler GCN approach to perform well for our case study. We now introduce GAMES, an augmented version of the linear GCN autoencoder for energy systems with multiple time resolutions.

\subsection{GAMES}
\begin{figure*}
    \centering
    \includegraphics[width=\textwidth]{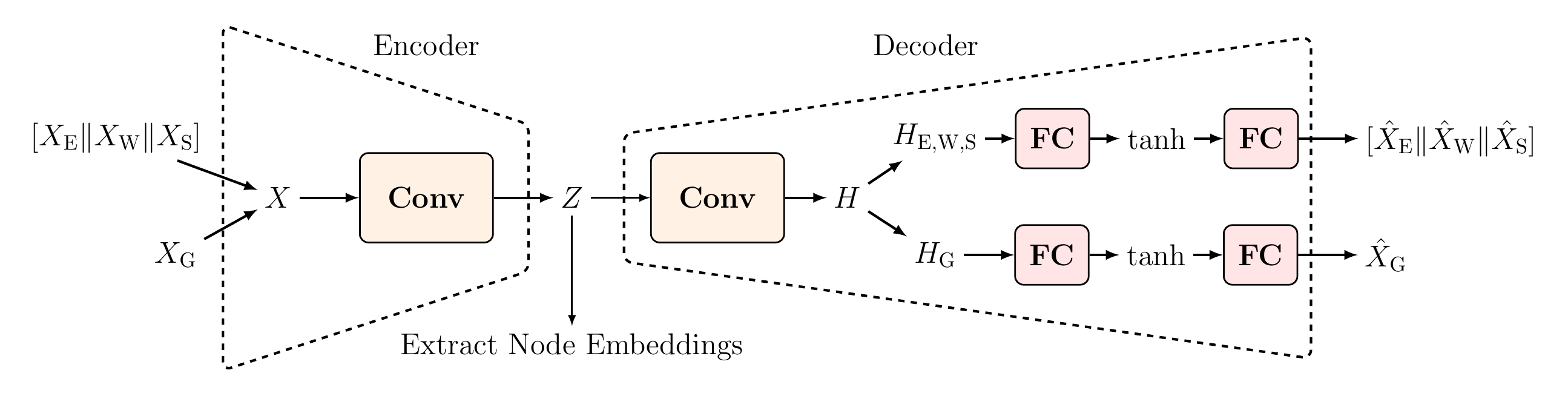}
    \caption{The GAMES Architecture. The electric power, wind CF, solar CF, and NG time series are combined into the block matrix $X$ with $n$ rows and $t_\mathrm{E}+t_\mathrm{W}+t_\mathrm{S}+t_\mathrm{G}$ channels. A single linear graph convolutional layer constructs matrix $Z$ by embedding each row of $X$ into $k$ dimensions. Another graph convolutional layer scales each row of $Z$ back to $t_\mathrm{E}+t_\mathrm{W}+t_\mathrm{S}+t_\mathrm{G}$ dimensions, which are then separated and fed through fully connected layers to reconstruct the two time series. After the model is trained, the embeddings are extracted by feeding the daily time series inputs through the encoder, at which point clustering is applied.}
    \label{fig:GAMES}
\end{figure*}

Our proposed GAMES architecture is designed as follows and illustrated in Fig. \ref{fig:GAMES}.

\subsubsection{Encoder}
Consider the power, wind CF, solar CF, and NG time series corresponding to day $i$, $X_\mathrm{E}^{(i)}$, $X_\mathrm{W}^{(i)}$, $X_\mathrm{S}^{(i)}$, $X_\mathrm{G}^{(i)}$. We begin by constructing the data matrix $X^{(i)}$ as
\begin{align*}
    X^{(i)} = 
    \begin{pmatrix}
 X_\mathrm{E}^{(i)} & X_\mathrm{W}^{(i)} & X_\mathrm{W}^{(i)} & \mathbf{0} \\
 \mathbf{0} & \mathbf{0} & \mathbf{0} & X_\mathrm{G}^{(i)}
    \end{pmatrix}.
\end{align*}
Note that $X^{(i)} \in \mathbb{R}^{n \times t}$, where $n\coloneqq n_\mathrm{E}+n_\mathrm{G}$ and $t\coloneqq t_\mathrm{E} + t_\mathrm{W} + t_\mathrm{W} + t_\mathrm{G}$. This is because capacity factor data exists for all nodes in the power network and utilizes the same network topology. $X^{(i)}$ is then passed through a single convolutional layer to produce the low-dimensional embedding $Z^{(i)} \in \mathbb{R}^{n \times k}$. The hyperparameter $k$ defines the bottleneck of the autoencoder architecture (i.e. the dimension of each node embedding) and consequently the tradeoff between compression and reconstruction loss. In our case study, we find $k=3$ to show a sufficient performance for our application of identifying representative days.

\subsubsection{Decoder}
$Z^{(i)}$ is passed through a convolutional layer to produce the embedding $H^{(i)} \in \mathbb{R}^{n \times t}$. This reconstructed matrix is then split along the second dimension into two blocks: $H_\mathrm{E,W,S}^{(i)} \in \mathbb{R}^{(n_\mathrm{E}+n_\mathrm{W}+n_\mathrm{S}) \times t}$ and $H_\mathrm{G}^{(i)} \in \mathbb{R}^{n_\mathrm{G} \times t}$. Each block is then passed to a separate series of fully connected layers with $\mathrm{tanh}$ activations that map the node embeddings in $H_\mathrm{E,W,S}^{(i)}$ and $H_\mathrm{G}^{(i)}$ respectively to the reconstructions $\hat{X}_\mathrm{E,W,S}^{(i)}$ and $\hat{X}_\mathrm{G}^{(i)}$. Finally, the tensor $\hat{X}_\mathrm{E,W,S}^{(i)}$ is split into the reconstructions $\hat{X}_\mathrm{E}^{(i)},\hat{X}_\mathrm{W}^{(i)},\hat{X}_\mathrm{S}^{(i)}$.

\subsection{Clustering}
After the model is trained, the power and NG time series from each day $i$, i.e. $(X_\mathrm{E}^{(1)},X_\mathrm{W}^{(1)},X_\mathrm{S}^{(1)},X_\mathrm{G}^{(1)}),\dots,(X_\mathrm{E}^{(N)},X_\mathrm{W}^{(N)},X_\mathrm{S}^{(N)},X_\mathrm{G}^{(N)})$, is passed through the encoder to generate the embedding matrices $Z^{(1)},\dots,Z^{(N)}$. Then, k-medoids clustering is applied to select a set of $K$ cluster medians, denoted by $\mathcal{S} \subset \{1,\dots,N\}$, and assign each day $i$ to a corresponding cluster $j \in \mathcal{S}$. We denote the set of days assigned to the cluster defined by day $j$ as $\mathcal{C}_j$. Given the number of clusters $K$, the k-medoids algorithm aims to minimize the objective function
\begin{align}\label{eq:cluster}
    \min \sum_{j \in \mathcal{S}} \sum_{i \in \mathcal{C}_j} \|Z^{(i)} - Z^{(j)}\|_F^2
\end{align}
\cite{hastie2001}. Note that every day in the dataset must be assigned to exactly one cluster. Semantically, (\ref{eq:cluster}) can be understood as aiming to ensure that the set of representative days $\mathcal{S}$ proportionately partitions the full set of days in the dataset by minimizing squared Euclidean distances in the latent space as constructed by the autoencoder.

\section{Capacity Expansion Model}
The result of the clustering algorithm is used to solve the CEM for joint power and NG planning, which is formulated as a GTEP. The problem determines the minimum investment cost and operational decisions for the year 2050 under various investment, operational, and policy constraints. The investment decisions include establishing new power plants, transmission lines, and pipelines as well as decommissioning existing plants. The operational constraints include minimum production, ramping, energy balance, transmission, and storage. We consider emission limits and minimum share of VREs as policy constraints. Importantly, in our formulation, the emissions constraint limits CO$_2$ emissions incurred by the consumption of NG in \emph{both} networks. 

We introduce our model with simplified notation in this section and provide a detailed formulation in the supplementary material \cite{suppMat}. Let $\ze = (\xe,\ye,\pe)$ represent the set of variables for the power system. The integer variable $\xe$ is the variable establishing plants, decommissioning plants, and establishing new transmission lines. The continuous variable $\pe$ captures the power generation in NG-fired plants while $\ye$ is a continuous variable that captures all the remaining variables including power generation from non NG-fired plants and power flow between nodes, storage, and load shedding variables. We use $\zg = (\xg,\yg,\fg)$ to denote the set of variables associated with the NG system. The mixed-integer variable $\xg$ is the set of all investment, storage, and load shedding decisions. The continuous variable $\yg$ represents the intra-network flow, i.e. the flow between NG nodes or the flow between NG nodes and NG storage facilities. The flow between NG and electricity systems is denoted by $\fg$. We formulate the joint power-NG system as follows:
\begin{subequations}\label{model:cc}
\begin{align}
\min \ &  (\CEone \xe +\CEtwo \ye+\CEthree \pe) + (\CGone \xg + \CGtwo  \yg+ \CGthree \fg)&\hspace{-0.5cm} \label{obj}\\
\text{s.t.} \ & \Ae \xe+\Be\ye  +\De \pe \leq \bEone& \label{e-c1}\\
    & \He \ye \geq \bEtwo& \label{e-c2}\\
    & \Ag \xg + \Bg \yg + \Dg \fg \leq \bGone& \label{ng-c1}\\
    &  \fg = \Eone \pe 
   &\label{coup1}\\
    &\Gtwo \yg + \Etwo \pe  \leq \eta&\label{coup2} \\
    &\xe \in \mathbb{Z}^+, \ye, \xg \in \mathbb{Z}^+\times \mathbb{R}^+, \pe, \yg,  \fg\in \mathbb{R}^+ & \label{e-c4}
\end{align}
\end{subequations}
The objective function \eqref{obj} minimizes the investment and operational costs for the power system (first term) and NG system (second term). The constraint~\eqref{e-c1} represents all investment, commitment, and operational constraints for the power system including the production limit, ramping, storage, and energy balance constraints. The constraint~\eqref{e-c2} enforces policy considerations such as the minimum requirement for renewable portfolio standard (RPS). The NG constraints are reflected in constraint~\eqref{ng-c1}, which includes technological and operational constraints such as the supply limit at each node, flow between NG nodes, and storage.

The coupling constraint~\eqref{coup1} ensures that NG-fired plants operate based on the gas flow they receive from the NG network. The second coupling constraint~\eqref{coup2} is the decarbonization constraint that limits emissions resulting from NG consumption to serve both electricity (via NG power plants) and non-power related NG loads to $\eta$. The coefficient matrices $\Eone$, $\Gtwo$, and $\Etwo$ represent the heat rate, emission factors for NG usage, and emission factor for NG-fired plants, respectively. Indeed, emissions from coal-fired plants is a major driver for decarbonization efforts and NG remains as primary fuel for which emissions need to be regulated. Therefore, given the declining role of coal in the US energy system, the constraint \eqref{coup2} reflects a futuristic setting where such plants are already decommissioned. 



\section{Input Data}

Using publicly available data, we consider the New England region and construct its corresponding power and NG network. We then calibrate the resulting networks using historical data. The power network consists of 188 nodes with 338 existing and candidate transmission lines. The NG network consists of 18 NG nodes and 7 storage nodes. We assume that each NG node is connected to two other storage nodes. We also assume that each power node is connected to three of its closest NG nodes. The Supplementary Information provides the details of the input data for the joint power-NG planning model \cite{suppMat}.

\section{Computational Experiments}
\subsection{GAMES Performance}
We train GAMES on a dataset of 292 days using the Adam optimizer with a learning rate of $0.001$. We use the full batch of 292 data points for each update step and perform early stopping to end training when the validation loss no longer decreases. We report the validation reconstruction loss on a set of 73 days for various node embedding dimensions $k$ in Table $\ref{tab:mse}$.
\begin{table}[h!]
\centering
\begin{tabular}{c|cccc}
\hline
\\ [-0.8em]
Embed. Dim.& $k=1$  & $k=2$  & $k=3$  & $k=4$  \\ \hline
\\ [-0.7em]
MSE Loss & 0.727 & 0.398 & 0.244 & 0.160 \\ \hline
\end{tabular}
\caption{The reconstruction loss shows diminishing returns for $k>3$ node embedding dimensions.}
\label{tab:mse}
\end{table}
We observe slightly diminishing returns for the validation reconstruction loss for $k>3$. Consequently, we proceed with our representative day selection using embeddings generated by the model corresponding to $k=3$.

\subsection{Representative Days Comparison}

\subsubsection{Setup}
\begin{figure*}[ht]
    \centering
    \begin{subfigure}[b]{0.46\textwidth}
        \centering
        \includegraphics[width=\textwidth]{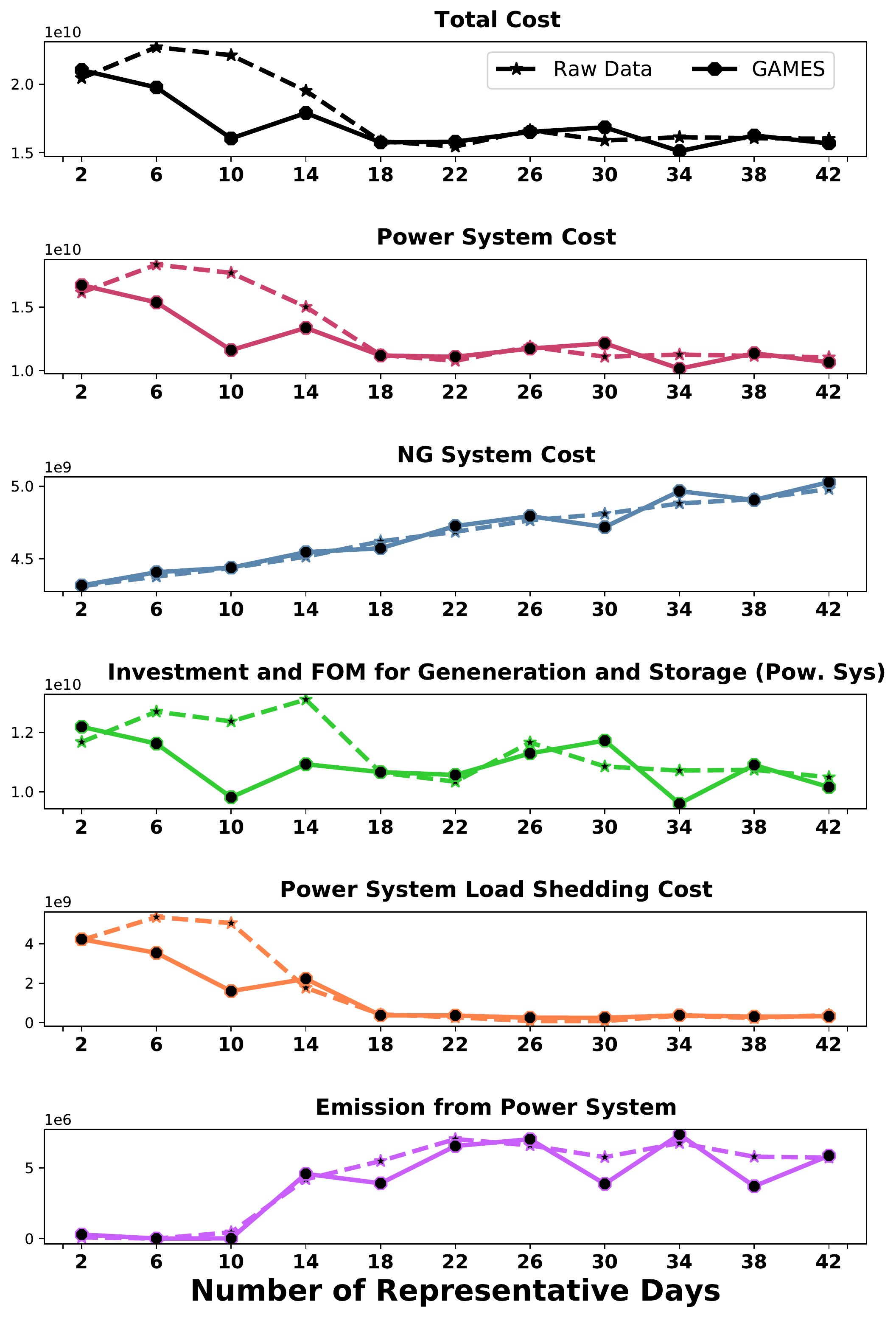}
        \caption{GAMES vs. raw data clustering comparison under an 80\% carbon reduction goal.}
        \label{fig:costs80}
    \end{subfigure}
    \hspace{4mm}
    \begin{subfigure}[b]{0.46\textwidth}
        \centering
        \includegraphics[width=\textwidth]{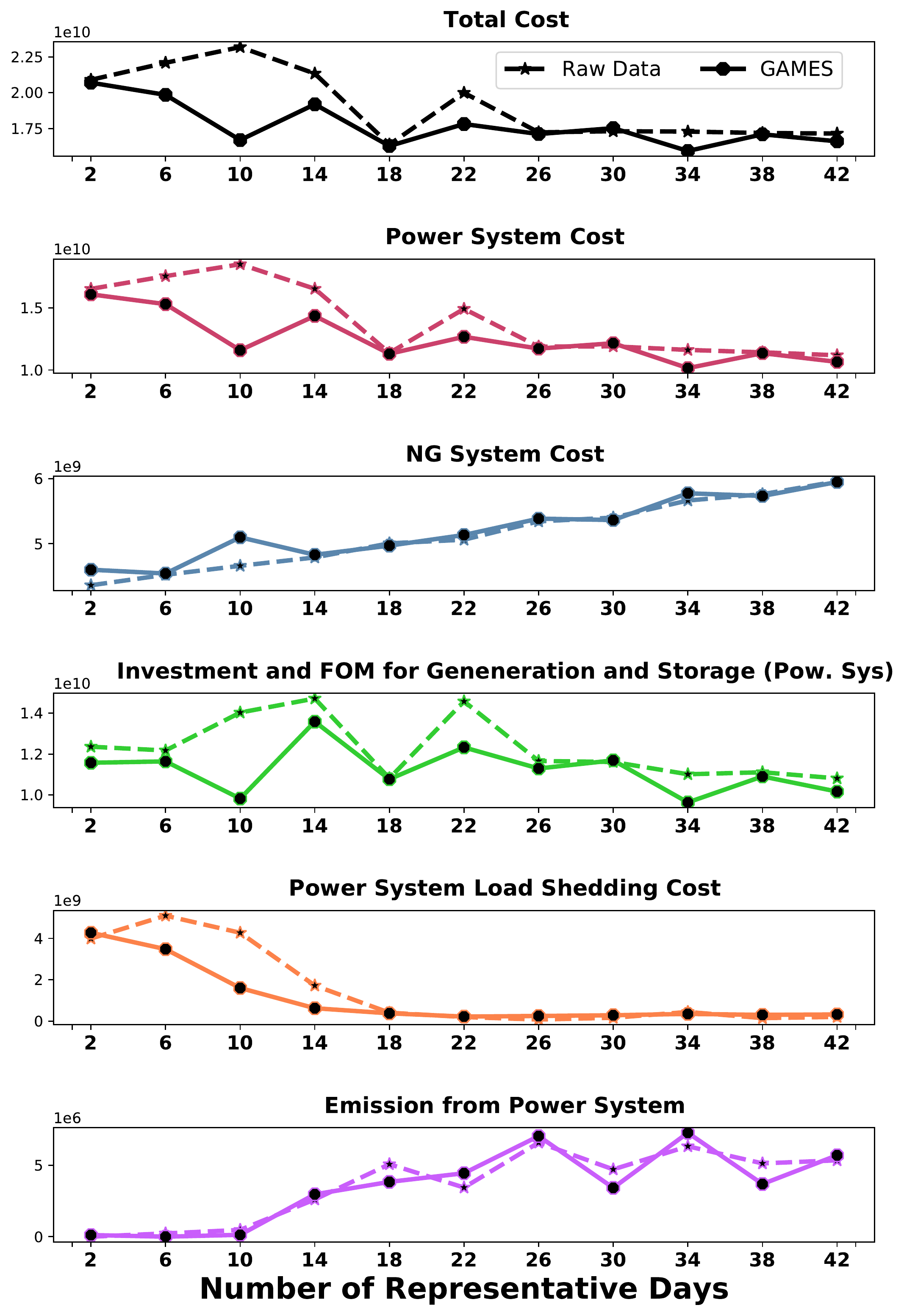}
        \caption{GAMES vs. raw data clustering comparison under a 95\% carbon reduction goal.}
        \label{fig:costs95}
    \end{subfigure}
    \caption{Various costs and power emission for different number of representative days under different decarbonization goals.}
\end{figure*}

We use the k-medoids clustering algorithm to obtain different sets of representative days. We apply the clustering algorithm to both raw data and the embeddings obtained from the GAMES model to compare the results of the proposed model. Accordingly, two different sets are obtained for each number of representative days. The optimization model over the full power network is prohibitively challenging even for a very small number of days. Therefore, we aggregate all buses in each state of the New England region to obtain a 6-node power network. This aggregation allows us to run the formulation for up to 42 representative days.  

We obtain a feasible solution in two steps for each set of representative days: (1) The optimization model is aggregated to the set of representative days for tractability and then solved. (2) Next, we consider the full planning horizon (the entire year of 2050) and set the integer decision variables (i.e. investment decisions) to the values determined in the first step. We note that the investment decision variables in our formulation are (a) the only integer-valued decision variables and (b) independent of planning periods. Therefore, fixing them reduces the remaining operational problem to a linear program (LP), which can be solved considerably faster. The resulting solution from the second step is a feasible solution to the full-year problem, with which we can analyze resulting costs and decisions.

\begin{table*}[hbtp]
\centering
\caption{Average percentage change when using GAMES approach for for various costs and power emissions.}
\label{tab:gap}
\begin{tabular}{ccccccc}
\toprule
Reduction Goal & Total & Power & NG    & Inv-FOM (Power)& Shedding  cost (Power) & Emission from Power Sys \\
\midrule
80$\%$                              & -5.14  & -7.03  & 0.24 & -4.64                 & -24.13         & -9.87           \\
95$\%$                              & -7.27  & -10.50 & 1.5 & -8.51                 & -27.80 &-3.31         \\
\bottomrule
\end{tabular}
\end{table*}

In our computational experiments, we consider two decarbonization goals of 80$\%$ and 95$\%$ where the former is the projected target for New England states \cite{Brattle2019}, and the latter aims reflects a radical decarbonization goal. 
Figures~\ref{fig:costs80} and \ref{fig:costs95} show the results under 80$\%$ and 95$\%$ emission reduction goals respectively. Both figures evaluate the following quantities for the clusters obtained from GAMES and raw data: i) ``Total Cost'' which is the objective function of model~\ref{model:cc}; ii) ``Power System Cost'' which is the first term in the objective function \eqref{obj}; iii) ``NG System Cost'' which is the second term in the objective function \eqref{obj}; iv) ``Investment and FOM for Generation and Storage (Pow. Sysm)'' (investment-FOM) which is part of the power system cost and captures the capital investment and fixed operating and maintenance (FOM) costs of installing new power plants and storage systems;  v) ``Power System Load Shedding Cost'' which is part of the power system cost and reflects the cost of unsatisfied electricity demand; and v) ``Emission from Power System'' which is the tonnage of emission as a result of operating NG-fired power plants in the power system. 
We use ``GAMES'' to denote the feasible solution for the set of days obtained by GAMES. We do not report the wall-clock times, but all instances are solved under 5 hours. As expected, run-times vary significantly depending on the number of representative days utilized; instances with 2 representative days typically run in fewer than 350 seconds, whereas 30-day instances may need to run for 2800 seconds. All instances are implemented in Python using Gurobi 9.5 and are run on the MIT Supercloud system with an Intel Xeon Platinum 8260 processor containing up to 96 cores
and 192 GB of RAM \cite{Supercloud2018}.

\subsubsection{Results}

Table~\ref{tab:gap} presents the percentage change in various quantities yielded by the GAMES representative days solution as compared to the solution using representative days selected from clustering the raw data. The cost comparisons are also plotted in Figures~\ref{fig:costs80} and \ref{fig:costs95}. We observe, on average, a 5.14$\%$ and 7.27$\%$ improvement (decrease) in the total cost when using GAMES under 80$\%$ and 95$\%$ decarbonization goals, respectively. This improvement may be attributed to GAMES' ability to model dependencies between power and NG system data. Under more stringent decarbonization targets, the share of VRE increases and the role of dispatchable power plants, such as NG-fired plants, diminishes. As a result, modeling the influence of capacity factors and their interactions with power and gas demands becomes more essential. This phenomenon may underlie our observation for the 22-day instance in which, while both approaches provide similar results under the 80$\%$ decarbonization goal, GAMES significantly outperforms the raw data clustering as measured by total cost for the higher decarbonization goal. As shown in Figures~\ref{fig:costs80} and \ref{fig:costs95}, the total cost from GAMES outperforms or matches the performance of the raw data clustering in all instances (except the 30-day instance under an 80$\%$ reduction goal). Interestingly, this disparity in performance is most drastic when 15 or fewer representative days are utilized under both decarbonization goals. This is worth noting as the optimization model instantiated on the full network topology (i.e. without aggregating nodes by state) is only tractable over a small set of representative days (i.e. after applying a very coarse temporal aggregation). It is especially important when the a model-year model only affords to consider a handful of representative days for each year. 


The power system cost largely drives variation in the total cost under both decarbonization goals -- the total cost is lower for all solutions with a lower power system cost. Note that the difference in performance is more pronounced in the power system cost compared to the total cost as indicated by the 7.03$\%$ and 10.50$\%$ power system cost improvement for GAMES under the 80$\%$ and 95$\%$ decarbonization goals. In Figure~\ref{fig:costs80}, this trend aligns with load shedding costs except for the 14-day instance. However, as the 24.13$\%$ decrease shows, the GAMES approach results in significantly lower load shedding on average. The 27.80$\%$ improvement in the load shedding cost for GAMES under the 95$\%$ goal is plotted in detail in Figure~\ref{fig:costs95}; GAMES outperforms the raw data clustering for all instances. Moreover, the GAMES approaches converges after 14 days with load shedding cost significantly lower than those instances utilizing fewer representative days.

In both figures the trends of investment-FOM cost and power system cost are the same, indicating that the power system cost is largely driven by investment-FOM cost, and to a lesser extent, by load shedding cost. This is expected as future energy systems will rely significantly on VREs such as solar and wind power, which only incur investment and FOM costs. Another interesting observation pertains to the quantity of emissions in the power system caused by operating NG-fired plants. 
Emissions for the power system are on average 9.87$\%$ and 3.31$\%$ lower for GAMES under the two decarbonization goals. This indicates a greater share of VREs in the GAMES approach, and correspondingly, a higher share of gas-fired plants in the raw data clustering approach. This is an interesting observation that may have significant implications for energy policy-making. In particular, it suggests that the results from the raw data clustering approach may be misleading as they do not sufficiently convey the radical changes required to transform the system from the current gas-dominant generation portfolio to a renewable-dominant power grid. 

NG system cost is another essential component of the total costs. Although NG costs are similar for GAMES and raw data clustering for each instance, the NG cost increases with the number of representative days. A possible explanation might be that neither GAMES nor raw data clustering aim to capture extreme days with separate clusters. Therefore, days with loads similar to extreme days are more likely to be selected as a cluster's medoid as the number of representative days increases, which inevitably raises the NG system cost. This consideration is also consistent with the observed load shedding cost for the power system, which is significantly higher for instances with fewer than 15 representative days, indicating that both approaches fail to account for extreme days in cluster medoids.

\section{Conclusion}
In this work, we propose GAMES, a graph convolutional autoencoder for modeling energy demand in interdependent electric power and  natural gas systems with heterogeneous nodes and different time resolutions. GAMES is able to exploit spatio-temporal demand patterns to learn efficient embeddings of interdependent power and NG networks. We apply the k-medoids clustering algorithm to these embeddings to identify a set of representative days with which we are able to tractably solve an energy system infrastructure planning problem calibrated for the joint power-NG system in New England. Our computational results show that the proposed framework outperforms clustering methods applied to the raw data and is effective in selecting a small number of representative days to provide high-quality feasible solutions for the optimization problem. 

The current work can be extended in multiple directions. The immediate extension of the GCN architecture is to explore alternative approaches to graph representation learning such as Laplacian sharpening \cite{park2019}. The extraction and inclusion of extreme days, or low-frequency days with unusually low or high demand is another potential next step which could prevent high load shedding costs and better represents the NG system's load patterns.

\bibliography{bibliography.bib}

\begin{thebibliography}{23}
\providecommand{\natexlab}[1]{#1}

\bibitem[{Barbar and Mallapragada(2022)}]{BarbarMallapragada2022}
Barbar, M.; and Mallapragada, D.~S. 2022.
\newblock Representative period selection for power system planning using
  autoencoder-based dimensionality reduction.
\newblock \emph{arXiv preprint arXiv:2204.13608}.

\bibitem[{(EIA)(2022)}]{eiaWebsite2021}
(EIA), E. I.~A. 2022.
\newblock EIA Website.
\newblock Website.
\newblock Accessed: 2022-2-18.

\bibitem[{Gielen et~al.(2019)Gielen, Gorini, Wagner, Leme, Gutierrez, Prakash,
  Asmelash, Janeiro, Gallina, Vale et~al.}]{GielenEtal2019}
Gielen, D.; Gorini, R.; Wagner, N.; Leme, R.; Gutierrez, L.; Prakash, G.;
  Asmelash, E.; Janeiro, L.; Gallina, G.; Vale, G.; et~al. 2019.
\newblock Global energy transformation: a roadmap to 2050.

\bibitem[{Hastie, Tibshirani, and Friedman(2001)}]{hastie2001}
Hastie, T.; Tibshirani, R.; and Friedman, J. 2001.
\newblock \emph{The Elements of Statistical Learning}.
\newblock Springer Series in Statistics. New York, NY, USA: Springer New York
  Inc.

\bibitem[{He et~al.(2018)He, Zhang, Liu, Wu, and
  Shahidehpour}]{HeEtal-survey2018}
He, C.; Zhang, X.; Liu, T.; Wu, L.; and Shahidehpour, M. 2018.
\newblock Coordination of interdependent electricity grid and natural gas
  network—a review.
\newblock \emph{Current Sustainable/Renewable Energy Reports}, 5(1): 23--36.

\bibitem[{Hoffmann et~al.(2020)Hoffmann, Kotzur, Stolten, and
  Robinius}]{HoffmannEtal2020-survey}
Hoffmann, M.; Kotzur, L.; Stolten, D.; and Robinius, M. 2020.
\newblock A review on time series aggregation methods for energy system models.
\newblock \emph{Energies}, 13(3): 641.

\bibitem[{Kipf and Welling(2017)}]{kipf2017}
Kipf, T.~N.; and Welling, M. 2017.
\newblock Semi-Supervised Classification with Graph Convolutional Networks.
\newblock In \emph{5th International Conference on Learning Representations,
  {ICLR} 2017, Toulon, France, April 24-26, 2017, Conference Track
  Proceedings}. OpenReview.net.

\bibitem[{Li et~al.(2022{\natexlab{a}})Li, Conejo, Liu, Omell, Siirola, and
  Grossmann}]{LiEtal2022-EJOR}
Li, C.; Conejo, A.~J.; Liu, P.; Omell, B.~P.; Siirola, J.~D.; and Grossmann,
  I.~E. 2022{\natexlab{a}}.
\newblock Mixed-integer linear programming models and algorithms for generation
  and transmission expansion planning of power systems.
\newblock \emph{European Journal of Operational Research}, 297(3): 1071--1082.

\bibitem[{Li et~al.(2022{\natexlab{b}})Li, Conejo, Siirola, and
  Grossmann}]{LiEtal2022}
Li, C.; Conejo, A.~J.; Siirola, J.~D.; and Grossmann, I.~E. 2022{\natexlab{b}}.
\newblock On representative day selection for capacity expansion planning of
  power systems under extreme operating conditions.
\newblock \emph{International Journal of Electrical Power \& Energy Systems},
  137: 107697.

\bibitem[{Li, Han, and Wu(2018)}]{li2018}
Li, Q.; Han, Z.; and Wu, X.-M. 2018.
\newblock Deeper Insights into Graph Convolutional Networks for Semi-Supervised
  Learning.
\newblock In \emph{Proceedings of the Thirty-Second AAAI Conference on
  Artificial Intelligence and Thirtieth Innovative Applications of Artificial
  Intelligence Conference and Eighth AAAI Symposium on Educational Advances in
  Artificial Intelligence}, AAAI'18/IAAI'18/EAAI'18. AAAI Press.
\newblock ISBN 978-1-57735-800-8.

\bibitem[{Liu, Sioshansi, and Conejo(2017)}]{LiuEtal2017}
Liu, Y.; Sioshansi, R.; and Conejo, A.~J. 2017.
\newblock Hierarchical clustering to find representative operating periods for
  capacity-expansion modeling.
\newblock \emph{IEEE Transactions on Power Systems}, 33(3): 3029--3039.

\bibitem[{Mallapragada et~al.(2018)Mallapragada, Papageorgiou, Venkatesh, Lara,
  and Grossmann}]{MallapragadaEtal2018}
Mallapragada, D.~S.; Papageorgiou, D.~J.; Venkatesh, A.; Lara, C.~L.; and
  Grossmann, I.~E. 2018.
\newblock Impact of model resolution on scenario outcomes for electricity
  sector system expansion.
\newblock \emph{Energy}, 163: 1231--1244.

\bibitem[{Park et~al.(2019)Park, Lee, Chang, Lee, and Choi}]{park2019}
Park, J.; Lee, M.; Chang, H.~J.; Lee, K.; and Choi, J.~Y. 2019.
\newblock Symmetric Graph Convolutional Autoencoder for Unsupervised Graph
  Representation Learning.
\newblock \emph{2019 IEEE/CVF International Conference on Computer Vision
  (ICCV)}, 6518--6527.

\bibitem[{Parsons, Haque, and Liu(2004)}]{parsons2004}
Parsons, L.; Haque, E.; and Liu, H. 2004.
\newblock Subspace Clustering for High Dimensional Data: A Review.
\newblock \emph{SIGKDD Explor. Newsl.}, 6(1): 90–105.

\bibitem[{Reuther et~al.(2018)Reuther, Kepner, Byun, Samsi, Arcand, Bestor,
  Bergeron, Gadepally, Houle, Hubbell et~al.}]{Supercloud2018}
Reuther, A.; Kepner, J.; Byun, C.; Samsi, S.; Arcand, W.; Bestor, D.; Bergeron,
  B.; Gadepally, V.; Houle, M.; Hubbell, M.; et~al. 2018.
\newblock Interactive supercomputing on 40,000 cores for machine learning and
  data analysis.
\newblock In \emph{2018 IEEE High Performance extreme Computing Conference
  (HPEC)}, 1--6. IEEE.

\bibitem[{Salha, Hennequin, and Vazirgiannis(2019)}]{salha2019}
Salha, G.; Hennequin, R.; and Vazirgiannis, M. 2019.
\newblock Keep it simple: Graph autoencoders without graph convolutional
  networks.
\newblock \emph{arXiv preprint arXiv:1910.00942}.

\bibitem[{Scott et~al.(2019)Scott, Carvalho, Botterud, and
  Silva}]{ScottEtal2019}
Scott, I.~J.; Carvalho, P.~M.; Botterud, A.; and Silva, C.~A. 2019.
\newblock Clustering representative days for power systems generation expansion
  planning: Capturing the effects of variable renewables and energy storage.
\newblock \emph{Applied Energy}, 253: 113603.

\bibitem[{Shuman et~al.(2012)Shuman, Narang, Frossard, Ortega, and
  Vandergheynst}]{shuman2012}
Shuman, D.; Narang, S.~K.; Frossard, P.; Ortega, A.; and Vandergheynst, P.
  2012.
\newblock The Emerging Field of Signal Processing on Graphs: Extending
  High-Dimensional Data Analysis to Networks and Other Irregular Domains.
\newblock \emph{IEEE Signal Processing Magazine}, 30.

\bibitem[{SI(2022)}]{suppMat}
SI. 2022.
\newblock Supplementary material available at: https://shorturl.at/bkHOU.

\bibitem[{Taubin(1995)}]{taubin1995}
Taubin, G. 1995.
\newblock A Signal Processing Approach to Fair Surface Design.
\newblock In \emph{Proceedings of the 22nd Annual Conference on Computer
  Graphics and Interactive Techniques}, SIGGRAPH '95, 351–358. New York, NY,
  USA: Association for Computing Machinery.
\newblock ISBN 0897917014.

\bibitem[{Teichgraeber and Brandt(2019)}]{TeichgraeberBrandt2019}
Teichgraeber, H.; and Brandt, A.~R. 2019.
\newblock Clustering methods to find representative periods for the
  optimization of energy systems: An initial framework and comparison.
\newblock \emph{Applied energy}, 239: 1283--1293.

\bibitem[{UN-FCCC(2015)}]{ParisAgreement2015}
UN-FCCC. 2015.
\newblock Decision 1/CP. 21, Adoption of the Paris Agreement.
\newblock In \emph{Report of the Conference of the Parties on Its Twenty-First
  Session, Held in Paris from}, volume~30.

\bibitem[{Weiss and Hagerty(2019)}]{Brattle2019}
Weiss, J.; and Hagerty, J.~M. 2019.
\newblock Achieving 80$\%$ GHG Reduction in New England by 2050.

\end{thebibliography}
\end{document}